# An Efficient Edge Detection Technique by Two Dimensional Rectangular Cellular Automata

Jahangir Mohammed
PG Department of Physics
Utkal University
Bhubaneswar, India
jahangir.isi@gmail.com

Deepak Ranjan Nayak
Department of Computer Science
College of Engineering and Technology
Bhubaneswar, India
depakranjannayak@gmail.com

*Abstract*—**This paper proposes a new pattern of two dimensional cellular automata linear rules that are used for efficient edge detection of an image. Since cellular automata is inherently parallel in nature, it has produced desired output within a unit time interval. We have observed four linear rules among $2^9$ total linear rules of a rectangular cellular automata in adiabatic or reflexive boundary condition that produces an optimal result. These four rules are directly applied once to the images and produced edge detected output. We compare our results with the existing edge detection algorithms and found that our results shows better edge detection with an enhancement of edges.**

## I. Introduction

Cellular Automaton is the common and most simple model of parallel computation. CA model consists of a regular grid of cells which are uniformly connected to each other and each cell consists of a finite number of states. Here all the cells behave as masters. On successive time intervals, state of each cell is updated with the state of the neighborhood cells and a specific updating rule. All the state of the cells in the grid are updated simultaneously [1, 2]. Since its mechanism is parallel the updating of the state of the cells in the infinite grid is one unit. Hence time complexity for updating state of the cells is the least. One dimensional CA is simply a linear array of cells. Two dimensional CA consists of rectangular or hexagonal grid of cells. A CA with one central cell and four near neighborhood cells is called a von Neumann/Five neighborhood CA whereas a CA having one central cell and eight near neighborhood cells is called Moore/Nine neighborhood CA [3, 4].

Image is viewed as a two dimensional CA model with initial configuration in which each pixel is represented by a cell and the pixel value is represented by the state of the cell. So, any updating rule can be applied once in an image at a particular time and its intensity of the pixels change simultaneously in the successive time interval. Due to this kind of behavior of CA model influences a large application in image processing. Edge detection is one of the most important and common technique in image processing. In an image, we generally concentrate on objects rather than on the background of an image. Hence objects are important. Edges characterize boundaries of objects. These are significant local changes of intensity in an image [5, 6, 7]. Based on the above idea many algorithms have been developed for edge detection, however this is still a challenging and an unsolved problem. In this paper we have proposed four linear CA rules which show better result for edge detection than some traditional methods.

## II. Rectanular CA

The architecture of CA depends on the geometrical shape of the cells and the inter connections among the cells. The inter connections are in such a manner that the model is inherently parallel and hence all the cells are updated simultaneously with successive interval of time. If the square cells are arranged in a finite dimension of a rectangle which is shown in Fig. 1, we called it as a rectangular CA. Generally CA depends on the dimension of the model, number of finite states of cells (*S*), neighborhood cells (*N*) and its distance (*r*), boundary condition and transition functions or rules (*F*). In our rectangular CA model, $S = \{0, 1\}$, N = 9, r = 1, adiabatic or reflexive boundary condition and linear updating rules.

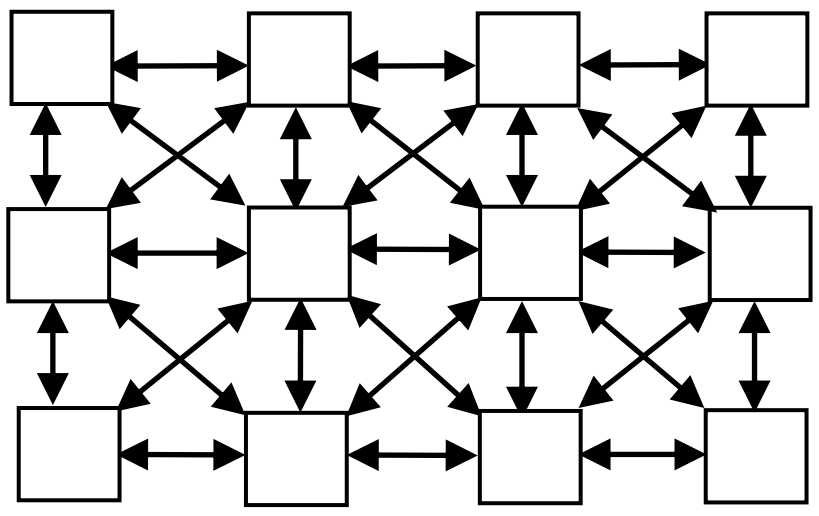

Figure 1. Rectangular CA model

### A. *Linear Rules*

For two states two dimensional CA with nine neighborhood, there are $2^{512}$ number of total rules among them $2^9$ number of rules are linear. 512 linear rules are determined by using nine basic rules showing in Fig 2. Taking XOR operation(/s) among basic rules, we get other 502 linear rules (excluding $Rule_0$) [2].

| | | |
|---|---|---|
| 64 | 128 | 256 |
| 32 | 1 | 2 |
| 16 | 8 | 4 |

Figure 2. Two Dimensional CA Linear Rules Convention.

The following example shows how a composite $Rule_{263}$ is calculated using basic rules and XOR operations.

*Example 1* $Rule_{263}$ is expressed in terms of basic rule matrices as follows:

$Rule_{263} = Rule_{256} \oplus Rule_4 \oplus Rule_2 \oplus Rule_1$.

### B. Boundary conditions

In fixed boundary condition the extreme cells are connected to logic 0/1 state. If the extreme cells are adjacent to each other then it is called periodic boundary condition. In adiabatic boundary condition extreme cells are replicate its state and in reflexive boundary condition mirror states replace the extreme cells. Showing an example, consider an one dimensional CA of length five and which has $x_1$, $x_2$, $x_3$, $x_4$ and $x_5$ states. The grey cells represent the neighborhood cells states of extreme cells. The following four boundary conditions are present.

- Fixed boundary condition.

| | | | | | | |
|---|---|---|---|---|---|---|
| 0/1 | $x_1$ | $x_2$ | $x_3$ | $x_4$ | $x_5$ | 0/1 |

- Periodic boundary condition.

| | | | | | | |
|---|---|---|---|---|---|---|
| $x_5$ | $x_1$ | $x_2$ | $x_3$ | $x_4$ | $x_5$ | $x_1$ |

- Adiabatic boundary condition.

| | | | | | | |
|---|---|---|---|---|---|---|
| $x_1$ | $x_1$ | $x_2$ | $x_3$ | $x_4$ | $x_5$ | $x_5$ |

- Reflexive boundary condition.

| | | | | | | |
|---|---|---|---|---|---|---|
| $x_2$ | $x_1$ | $x_2$ | $x_3$ | $x_4$ | $x_5$ | $x_4$ |

### C. Successor Matrix

If same rule is applied once to all the cells of a CA model then it is called uniform CA. At different time steps, the successor matrix is produced of a problem matrix after uniform rule is applied that is shown in the following example.

*Example 2* Shows how we get a successor matrix when uniform $Rule_{263}$ is applied once to the image $P_{m\times n}$.

$Rule_{263} = Rule_{256} \oplus Rule_4 \oplus Rule_2 \oplus Rule_1$

$P^{263} = P^{256} \oplus P^4 \oplus P^2 \oplus P^1$

where $P^1$ is the successor matrix after $Rule_1$ is applied once to P, $P^2$ is the successor matrix after $Rule_2$ is applied once to P and similarly for $P^4$ and $P^{256}$.

## III. PROPOSED METHOD

In this section we have discussed about the four linear rules in details and methodology to get desired output.

### A. Finding Linear Rules for Edge Detection

In Fig. 2 eight neighbor cells have four borders. Each boarder consists of three cells. For edge detection, the state of the central cell is updated with neighboring three border cells and the central cell. Now, eight cells can be divided into four borders (BLRT) as shown in Fig. 3, they are.

a) Bottom border ($Rule_{16}$, $Rule_8$ and $Rule_4$)

b) Left border ($Rule_{64}$, $Rule_{32}$ and $Rule_{16}$)

c) Right border ($Rule_{256}$, $Rule_2$ and $Rule_4$)

d) Top border ($Rule_{256}$, $Rule_{128}$ and $Rule_{64}$)

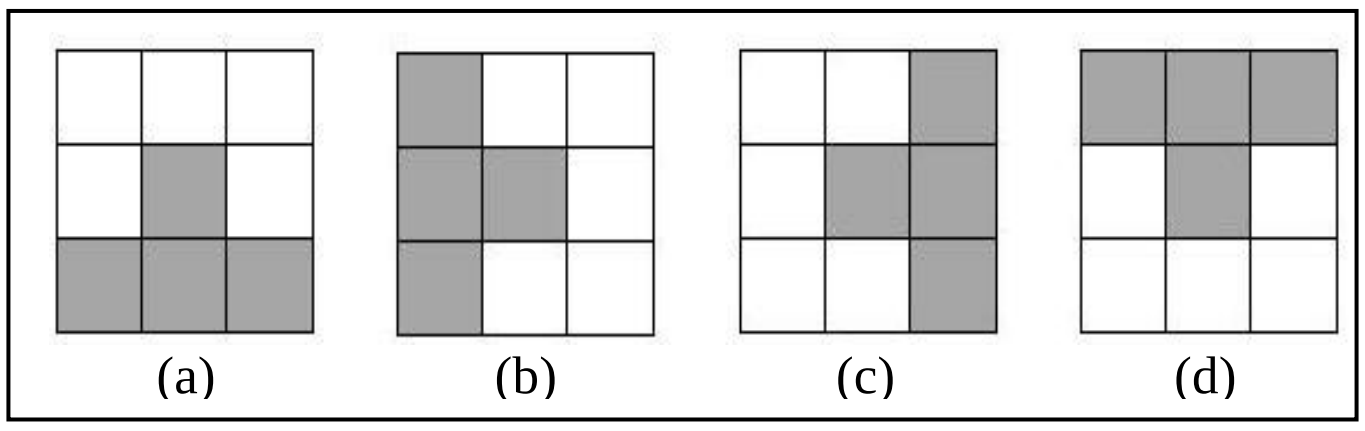

Figure 3. Proposed pattern of CA Rules: (a) Bottom border and a central cell is $Rule_{29}$, (b) Leftborder and a central cell is $Rule_{113}$, (c) Right border and a central cell is $Rule_{263}$, (d) Top border and a central cell is $Rule_{449}$.

Considering BLRT and a central cell, we have found $Rule_{29}$, $Rule_{113}$, $Rule_{263}$ and $Rule_{449}$ are used for edge detection. In the above figure, one can clearly observed that the pattern of the gray cells looks like English alphabet **T**.

The following examples demonstrate edge detection of an image using above rules.

*Example 3* Let us consider a 8×8 RCA whose initial configuration is an image $P_{8\times8}$. Now uniform $Rule_{29}$ applied once to the image $P_{8\times8}$ with adiabatic boundary condition and produced output $P^{29}_{8\times8}$.

$$P = \begin{pmatrix} 0&0&0&0&0&0&0&0 \\ 0&0&0&0&0&0&0&0 \\ 0&0&1&1&1&1&0&0 \\ 0&0&1&1&1&1&0&0 \\ 0&0&1&1&1&1&0&0 \\ 0&0&1&1&1&1&0&0 \\ 0&0&0&0&0&0&0&0 \\ 0&0&0&0&0&0&0&0 \end{pmatrix}_{8\times8} \xrightarrow{Rule_{29}} \begin{pmatrix} 0&0&0&0&0&0&0&0 \\ 0&1&0&1&1&0&1&0 \\ 0&1&1&0&0&1&1&0 \\ 0&1&1&0&0&1&1&0 \\ 0&1&1&0&0&1&1&0 \\ 0&0&1&1&1&1&0&0 \\ 0&0&0&0&0&0&0&0 \\ 0&0&0&0&0&0&0&0 \end{pmatrix} = P^{29}_{8\times8}.$$

*Example 4* Consider the same 8×8 RCA whose initial configuration is a binary image $P_{8\times8}$. Uniform $Rule_{449}$ applied once to the image $P_{8\times8}$ with reflexive boundary condition and produced output $P^{449}_{8\times8}$.

$$P = \begin{pmatrix} 0\,0\,0\,0\,0\,0\,0\,0 \\ 0\,0\,0\,0\,0\,0\,0\,0 \\ 0\,0\,1\,1\,1\,1\,0\,0 \\ 0\,0\,1\,1\,1\,1\,0\,0 \\ 0\,0\,1\,1\,1\,1\,0\,0 \\ 0\,0\,1\,1\,1\,1\,0\,0 \\ 0\,0\,0\,0\,0\,0\,0\,0 \\ 0\,0\,0\,0\,0\,0\,0\,0 \end{pmatrix}_{8\times8} \xrightarrow{Rule_{449}} \begin{pmatrix} 0\,0\,0\,0\,0\,0\,0\,0 \\ 0\,0\,0\,0\,0\,0\,0\,0 \\ 0\,0\,1\,1\,1\,1\,0\,0 \\ 0\,1\,1\,0\,0\,1\,1\,0 \\ 0\,1\,1\,0\,0\,1\,1\,0 \\ 0\,0\,1\,1\,1\,1\,0\,0 \\ 0\,1\,0\,1\,1\,0\,1\,0 \\ 0\,0\,0\,0\,0\,0\,0\,0 \end{pmatrix} = P^{449}_{8\times8}.$$

### B. *Methodology*

*Step I:* Convert any image to its corresponding binary image format by considering a suitable threshold value.

*Step II:* Apply one of the above four rules i.e., $Rule_{29}$, $Rule_{113}$, $Rule_{263}$ and $Rule_{449}$ to the matrix generated at *Step II* with adiabatic/reflexive boundary condition.
Finally, an edge detected image is produced.

## IV. Experimental Results

In this section, we have demonstrated the results of the proposed method on a gray scale blood cell image of size 256×249 and a gray scale cameraman image of size 256×256. Here, we have applied adiabatic boundary condition to the blood cell image and reflexive boundary condition to the cameraman image. The methods used here are implemented in MATLAB 2010a tool. Figures 4(a) and 5(a) illustrates the original gray images and figures 4(b) and 5(b) are their corresponding binary images which is done by taking an appropriate threshold value. The results of above four rules are shown in (c), (d), (e) and (f) in each figure.

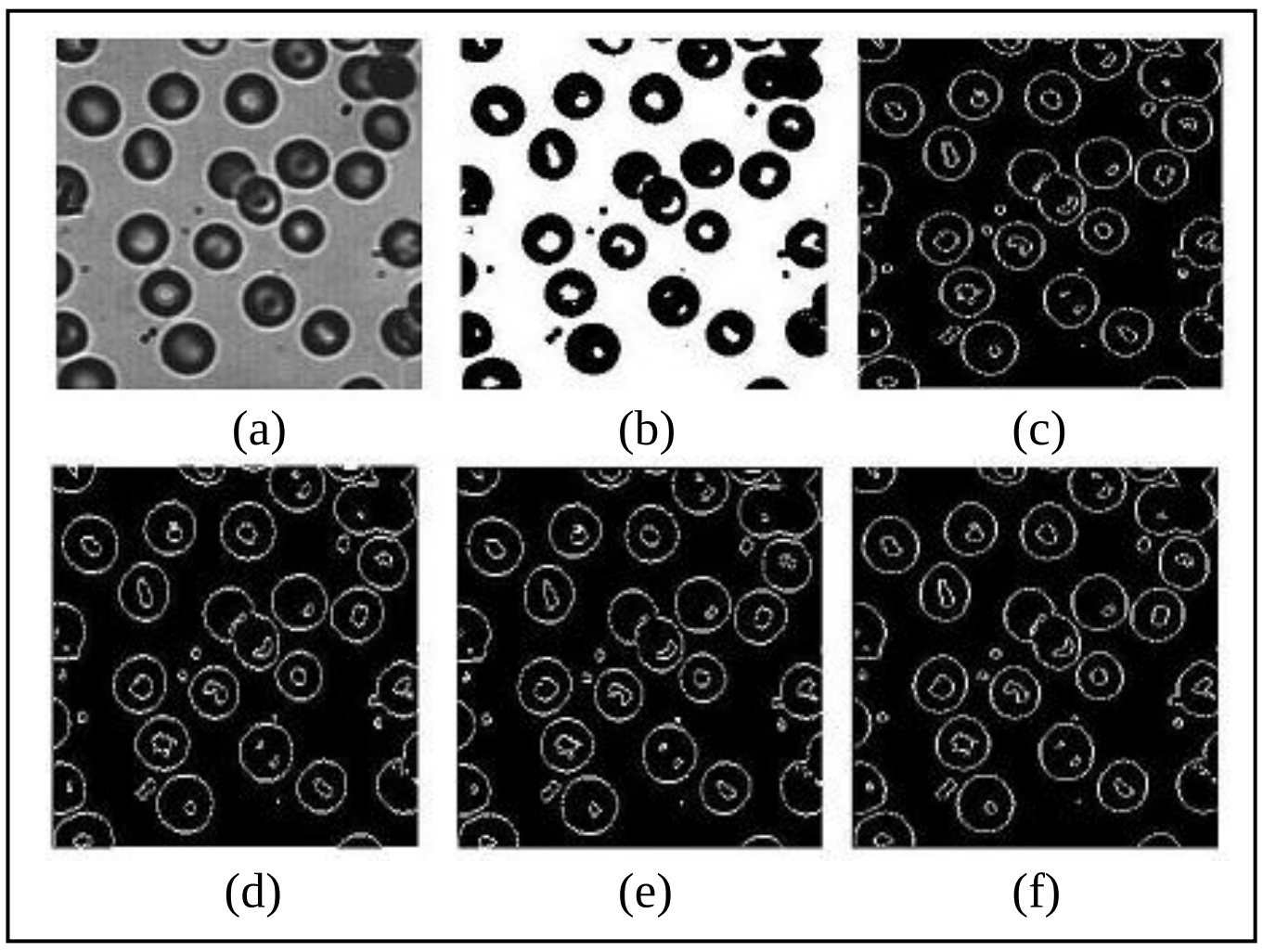

Figure 4. Edge detection of blood cell image based on proposed RCA rules with adiabatic boundary condition (a) Gray image, (b) ) Its binary Image, (c) $Rule_{29}$, (d) $Rule_{113}$, (e) $Rule_{263}$ and (f) $Rule_{449}$

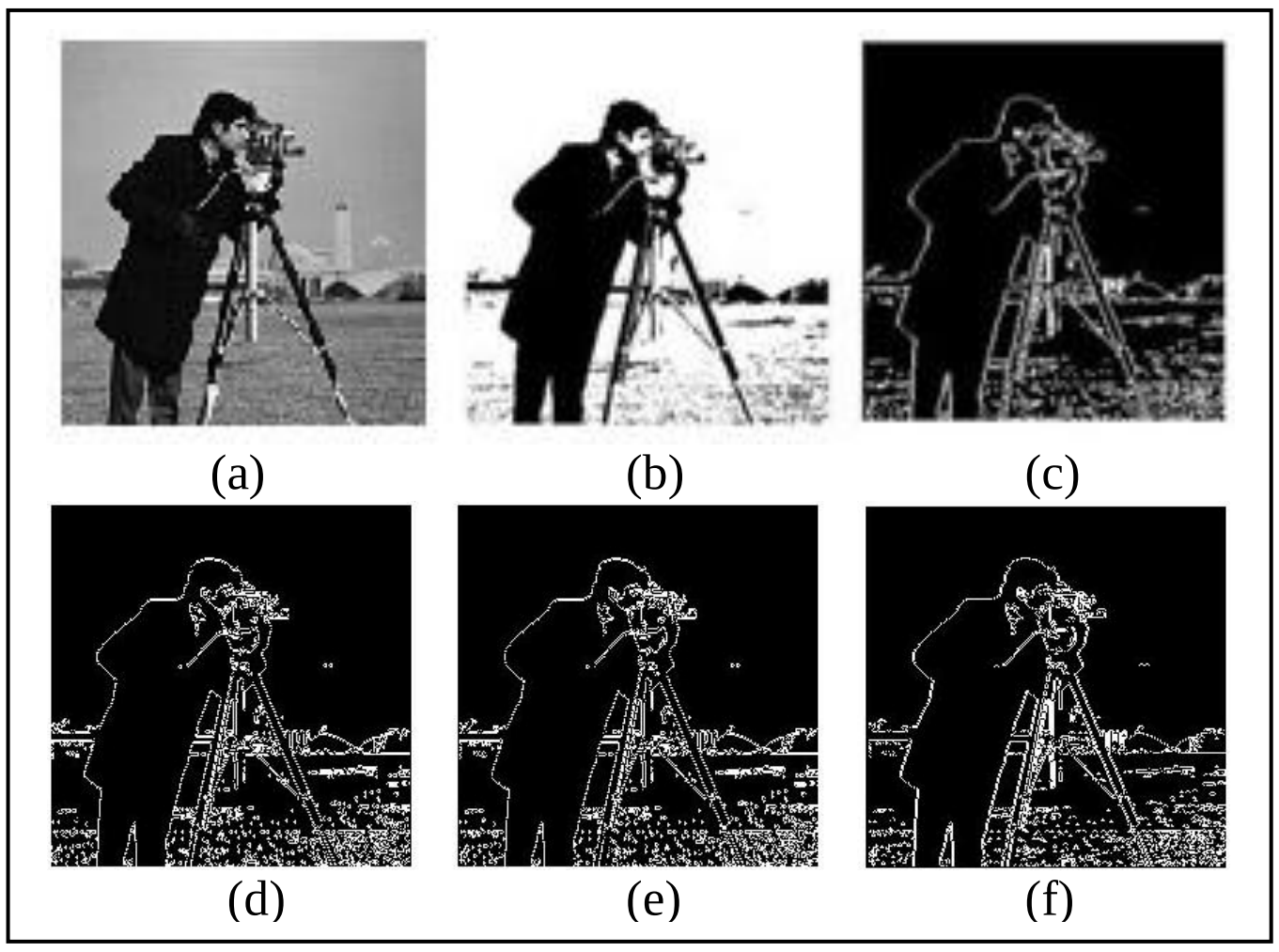

Figure 5. Edge detection of a cameraman image based on proposed RCA rules with reflexive boundary condition (a) Gray image, (b) ) Its binary Image, (c) $Rule_{29}$, (d) $Rule_{113}$, (e) $Rule_{263}$ and (f) $Rule_{449}$

## V. Comparision

In this subsection, we compare the experimental results of the proposed method with some of the traditional methods. There exist many algorithms such as Prewitt, Robert, Sobel, Log and Canny that can be used for edge detection but among them Canny produces optimal result in [7]. Here, we consider

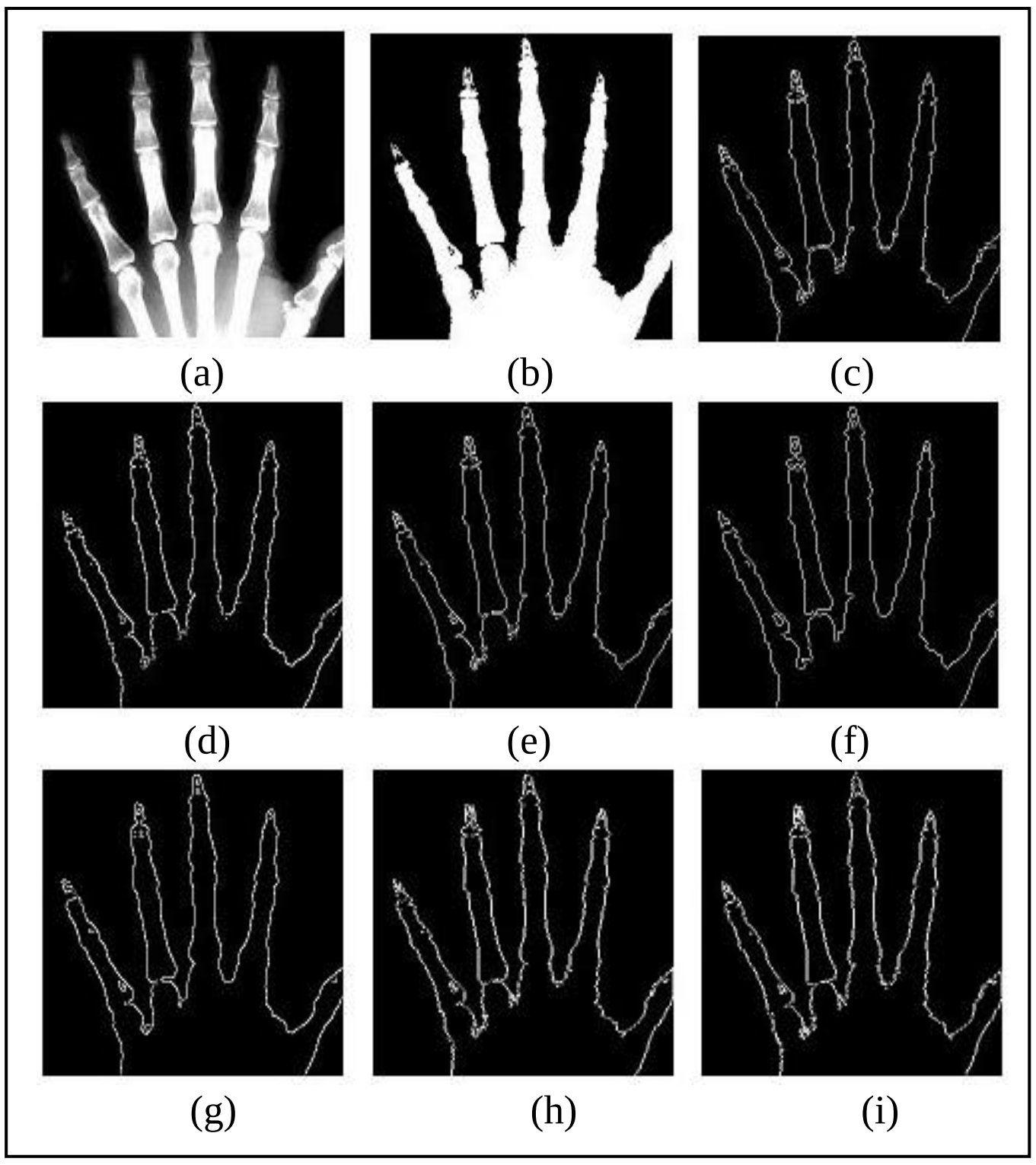

Figure 6. Edge detection of an x-ray image using different methods (a) Gray image, (b) ) Its binary Image, (c) Prewitt, (d) Robert, (e) Sobel, (f) LoG, (g) Canny, (h) $Rule_{29}$, and (i) $Rule_{449}$

a x-ray image of size 302×270 and apply the four optimal CA rules to its binary converted image. The results are compared in Fig. 6. From Fig. 6, we have found that the accuracy of edge detection using all the methods are not desirable in comparison to our proposed RCA rules. We have studied traditional methods by taking different images, in most cases Prewitt, Sobel, Robert and Log failed to detect an edge. But in case of Canny, the major disadvantage is smoothening out the rough edges. Whereas our four rules shows exact edge of an object with all its roughness. Canny method sometimes give false edges which is not present in the original iamge. But, due to the paucity of space we have present only one image for comparison in this section. From the visualization of above figures, one can demonstrates that the proposed method gives better edge detection of an image than all the existing methods in terms of contrast enhancement. That is the results are more suitable for further analysis.

## VI. Conclusion

Among 512 linear CA rules we have found only four linear rules which have better edge detection property. Each proposed rule have better edge detection with an enhancement of an image than present edge detection algorithms. Due to high contrast of an image the edges of objects shown clearly. Among all the traditional methods Canny edge detector yields better edge than other traditional detectors. We have found some cases our technique gives better result than Canny edge detection algorithm. RCA will reduced the time required for finding an edge of an image. Our proposed method also simple to understand and easy to implement. By combining these linear rules we can also study different image processing properties. And also one can studies behavior of hybrid CA in image processing. One of the big challenge of this field is finding some non-linear rules among $2^{512}$ - 512 non-linear rules that can solve a particular task.

## Acknowledgment

The authors wish to thank to Professor Swadheenanda Pattanayak (Former Director of Institute of Mathematics and Application), Professor Swapna Mahapatra (Professor in Physics, Utkal University), Professor Prashanta Kumar Patra (HOD CSE,College of Engineering and Technology) and Dr. Sudhakar Sahoo (Faculty in Computer Science, Institute of Mathematics and Application) for their useful discussions and constant encouragement.